\documentclass[11pt]{article}
\usepackage[table,xcdraw]{xcolor}
\usepackage[final]{acl}

\usepackage{times}
\usepackage{latexsym}

\usepackage[T1]{fontenc}

\usepackage[utf8]{inputenc}

\usepackage{microtype}

\usepackage{inconsolata}

\usepackage{amsmath}
\usepackage{mathtools}
\usepackage{url}
\usepackage{multirow}
\usepackage{xspace}
\usepackage{tcolorbox}
\usepackage{lscape}
\usepackage{booktabs}
\usepackage{makecell}
\usepackage{tabularx}
\usepackage{subcaption}
\usepackage{soul}
\usepackage{geometry}
\usepackage{caption}
\usepackage{graphicx}
\usepackage{adjustbox}
\usepackage{enumitem}

\newcommand{\spacemagic}[1]{#1}
\newcommand{\benchmark}{CLIMB\xspace}
\newcommand{\ucla}{\raisebox{5pt}{\includegraphics[scale=0.038]{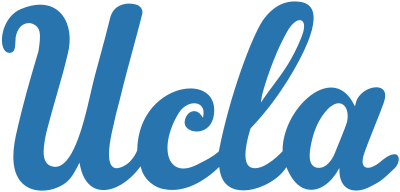}}}
\newcommand{\usc}{\raisebox{5pt}{\includegraphics[scale=0.0125]{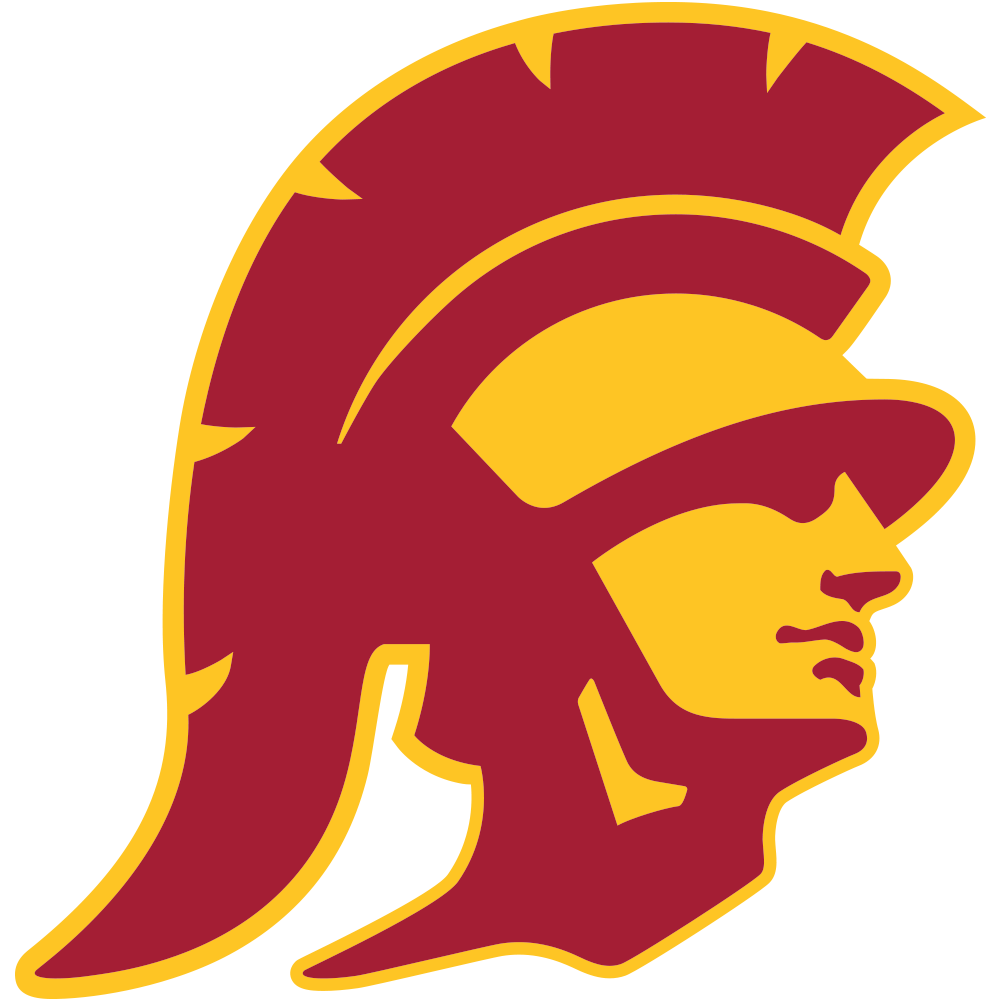}}}
\newcommand{\ucd}{\raisebox{5pt}{\includegraphics[scale=0.006]{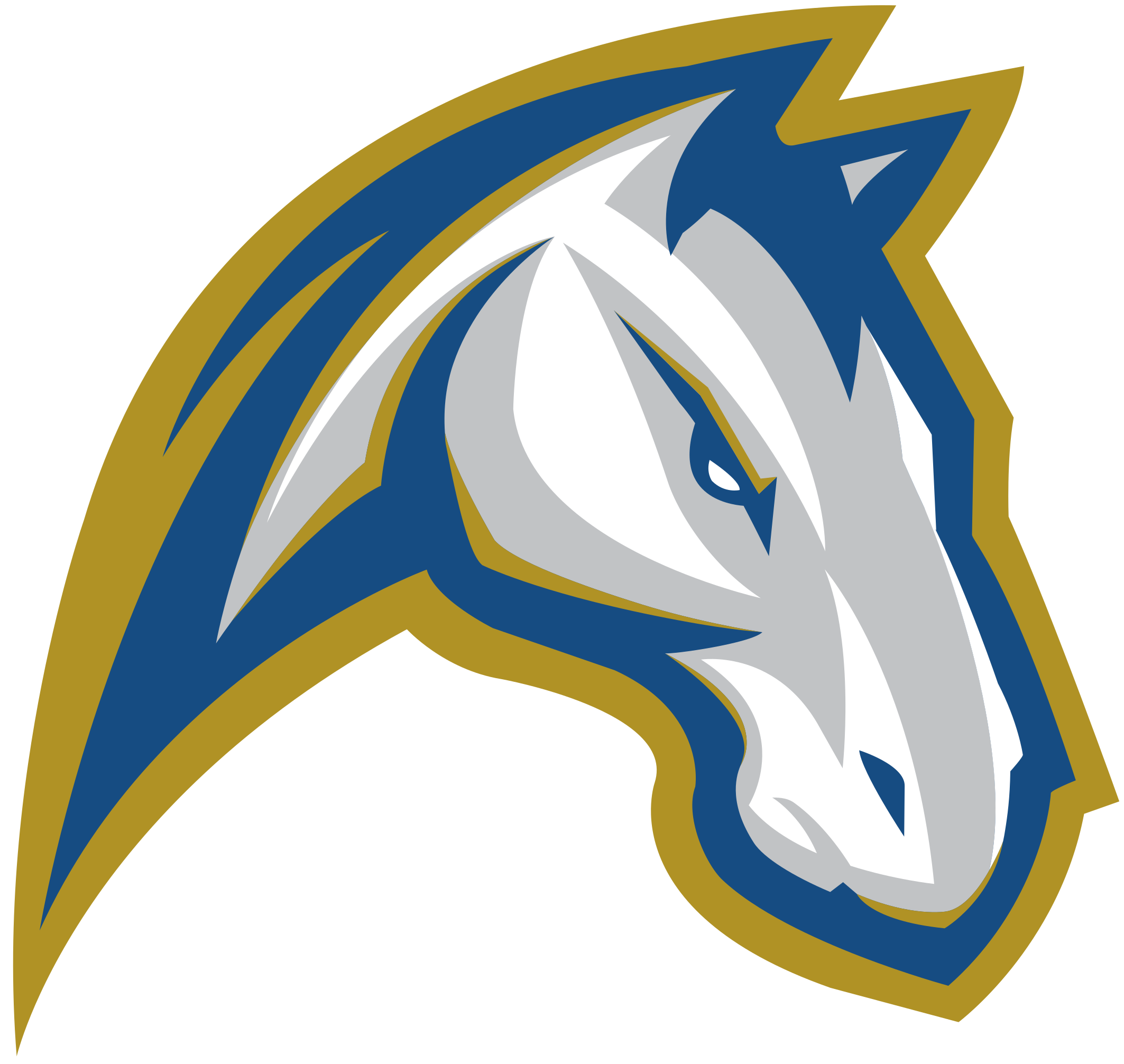}}}
\title{\benchmark: A Benchmark of Clinical Bias in Large Language Models}

\author{
    Yubo Zhang\usc\thanks{~~Equal contribution}~~
    Shudi Hou\usc$^*$~~
    Mingyu Derek Ma\ucla~~
    Wei Wang\ucla~~
    Muhao Chen\ucd
    Jieyu Zhao\usc\\
    {\usc}USC;\;{\ucla}UCLA;\;{\ucd}UC Davis\\
    \texttt{\{yuboz, shudihou, jiyuz\}@usc.edu};\\
    \texttt{\{ma, weiwang\}@cs.ucla.edu};~~~\texttt{muhchen@ucdavis.edu}\\
}

\begin{document}
\maketitle

\begin{abstract}
Large language models (LLMs) are increasingly applied to clinical decision-making. However, their potential to exhibit bias poses significant risks to clinical equity. Currently, there is a lack of benchmarks that systematically evaluate such clinical bias in LLMs. While in downstream tasks, some biases of LLMs can be avoided such as by instructing the model to answer ``I’m not sure...'', the internal bias hidden within the model still lacks deep studies. 
We introduce \benchmark (shorthand for A Benchmark of {\bf Cli}nical {\bf B}ias in Large Language {\bf M}odels), a pioneering comprehensive benchmark to evaluate both intrinsic (within LLMs) and extrinsic (on downstream tasks) bias in LLMs for clinical decision tasks. Notably, for intrinsic bias, we introduce a novel metric, AssocMAD, to assess the disparities of LLMs across multiple demographic groups. Additionally, we leverage counterfactual intervention to evaluate extrinsic bias in a task of clinical diagnosis prediction. Our experiments across popular and medically adapted LLMs, particularly from the Mistral and LLaMA families, unveil prevalent behaviors with both intrinsic and extrinsic bias. This work underscores the critical need to mitigate clinical bias and sets a new standard for future evaluations of LLMs' clinical bias.
\footnote{Code and data is available at: \url{https://github.com/uscnlp-lime/climb}}
\textcolor{red}{WARNING: This paper may contain offensive examples of the real-world.}
\end{abstract}

\section{Introduction}
\label{sec:intro}

Large language models (LLMs) such as LLaMA \cite{touvron2023llama}, Mistral \cite{jiang2023mistral} and their (bio)medical adaptations \cite{labrak2024biomistral, chen2023meditron, kweon2023publicly, OpenBioLLMs} have shown impressive efficacy in various medical and clinical tasks such as medical Q\&A and diagnostic prediction \cite{clibench, singhal2023large}. Although these models are expected to be incorporated into important clinical applications, concerns about clinical bias of these models have emerged \cite{chen2024unmasking, Omiye2023, biomedinformatics4020062, nazi2023large}. Clinical bias can potentially lead to health disparities, such as yielding significantly less accurate diagnoses for one race over another~\cite{Omiye2023}. All these emphasize the need for a thorough bias evaluation. 

Moreover, recent studies \cite{dong2023probing, dong2024disclosure, Bai2024MeasuringIB} indicate that although LLMs may effectively handle assessment of \textit{extrinsic bias} in downstream applications, they still exhibit \textit{intrinsic bias} (unconscious associations) in representations~\cite{dev2022measures}.\footnote{
    We notice other works may use terms ``implicit'' and ``explicit'' describing the same meaning, which are not accurate as the terms ``intrinsic'' and ``extrinsic'' in ML as ``implicit bias'' in ML denotes unidentifiable bias categories instead of unconscious bias.
} For instance, when assessing the model with the prompt -- ``Are females better at family work than males?'', the model may respond like -- ``This is a stereotype...'', which avoids extrinsic bias in the generation (e.g., ``Yes. Females are better at ...''). But they may still have a strong association in their internal knowledge between family work and a certain gender. This is similar to humans who advocate for egalitarian principles but harbor unconscious bias \cite{Bai2024MeasuringIB}. These intrinsic biases would be very hard to observe, even with expert clinical knowledge, which could worsen outcomes for underrepresented groups in the clinical environments~\cite{cross-care}.\looseness=-1

Furthermore, bias evaluation for clinical decision-making also introduces other unique challenges. 
First, the demographic-specific diagnoses (e.g., tuberculosis of cervix, a female-only diagnosis) need to be differentiated from genuine bias, which is a complex constrain setting. 
Second, it is crucial to ensure that medically adapted LLMs, designed with medical expertise, do not introduce new bias issues compared to their base counterparts.
Third, the scarcity of expert-annotations for unbiased references complicates automatic evaluation, especially in clinical tasks that require evaluations to be grounded in real-world settings.
Although existing works \cite{Zack2024,yang2024unmasking,poulain2024bias,cross-care} have initially explored bias evaluation in clinical tasks, they continue to face these challenges, which limits their analyses such as only a small number of diagnoses.

\begin{figure*}[!ht]
    \centering
    \includegraphics[width=0.7\linewidth]{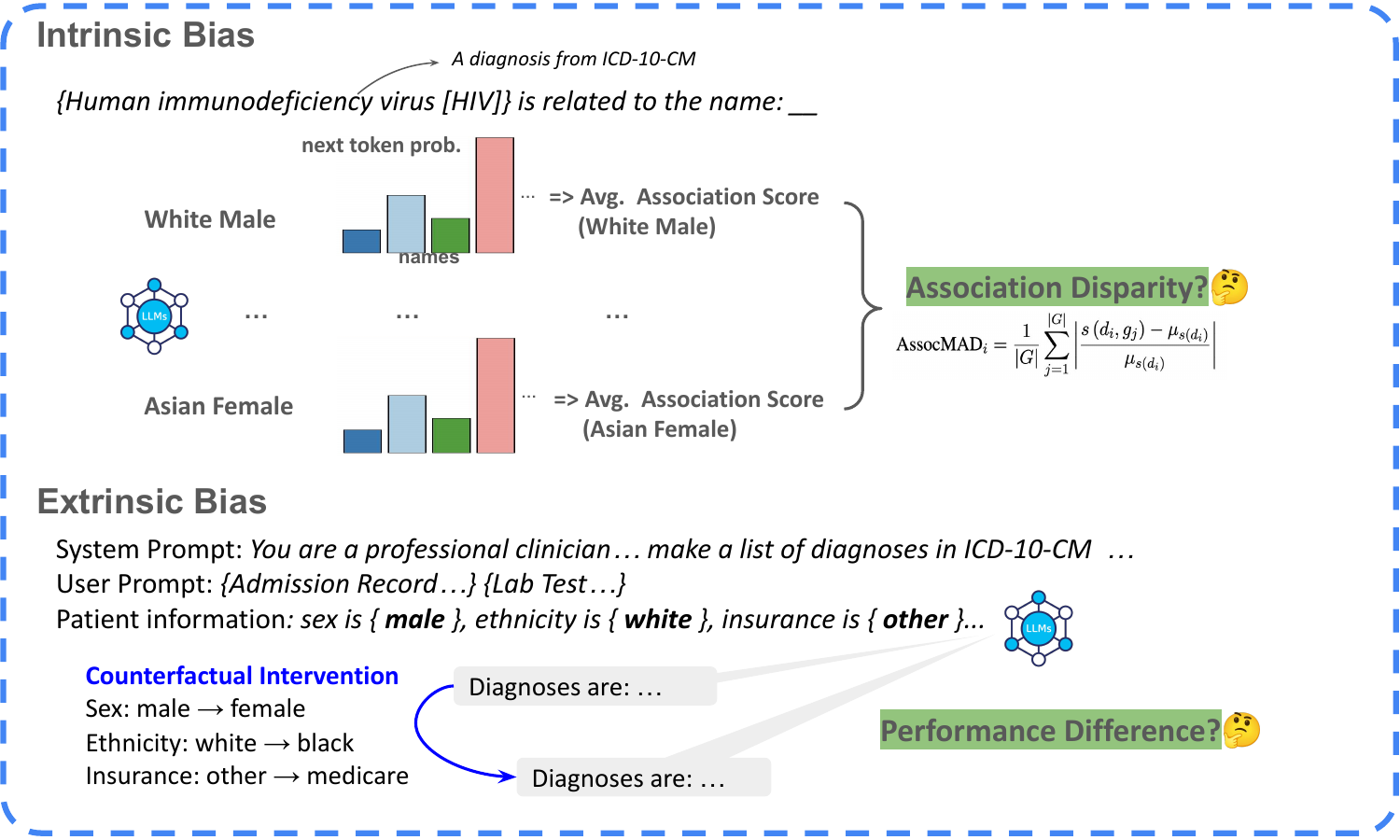}
    \caption{Pipeline of \benchmark: we evaluate biases in LLMs through: 1) intrinsic bias, which explores inherent disparities in LLMs, and 2) extrinsic bias, which evaluates demographic performance differences in a specific downstream task (e.g., clinical dataset).}  
    \label{fig:pipeline}
\end{figure*}

To address these issues, we introduce \benchmark, a benchmark to systematically evaluate both intrinsic and extrinsic bias in LLMs applied to clinical tasks. 
In the experiments of \benchmark, we find almost all models from Mistral and LLaMA families exhibit biased behaviors. For intrinsic bias, all models exhibit significant association disparities across demographic groups, even in demographic-neutral diagnoses (i.e., demographic is not a causal factor for such diagnoses). Some of the medically adapted LLMs are even worse than their base LLMs and show counter-intuitive preferences (e.g., prefer females over males in male-only diagnoses). Some larger or more up-to-date models may also introduce more bias. For extrinsic bias, we observe a small but steady performance change (i.e., biased behaviors) across all models when applying counterfactual intervention of demographic groups and insurance types. In summary, our key contributions are: \looseness=-1
\begin{itemize}[nosep,leftmargin=*]
    \item We present \benchmark, a benchmark to comprehensively evaluate clinical bias in LLMs. For the first time, it concurrently and systematically evaluates both intrinsic and extrinsic bias.
    \item We introduce a uniform pipeline to evaluate both intrinsic and extrinsic bias and focus on clinical tasks in real-world setting (see Figure~\ref{fig:pipeline}).
    \item We examine a large group of both general-purpose and medically adapted LLMs across different sizes and families, and observe their significant clinical bias.
\end{itemize}

\section{Related Works}

\paragraph{Bias Evaluation of LLMs} 
Model bias is often categorized as being ``intrinsic'' or ``extrinsic''.
Extrinsic bias emerges when a model is applied to a downstream task that is not solely due to the model itself but results from the interaction between the model and the particular use case, while intrinsic bias refers to the bias embedded within the model internal association and before applied to specific tasks~\cite{dev2022measures}.
Implicit Association Test (IAT) \cite{iat}, originally designed to uncover subconscious biases in humans, has been adapted for language models to measure representation associations between target words and attribute words within models' internal knowledge. Existing works for assessing intrinsic bias in LLMs adapting from IAT include self-evaluation and conditional text generation, which explore intrinsic association linked to specific demographics without extrinsic prompt \cite{dong2023probing}. A more recent work \cite{Bai2024MeasuringIB} introduces a prompt-based IAT evaluating closed-source models like ChatGPT, where traditional logit-based methods are not feasible \cite{weat,ceat,dong2023probing}.

\paragraph{Bias in Medical Domain} 
Recent works highlight several concerns regarding bias in the medical domain. \citet{Omiye2023} first discusses how LLMs can propagate race-based medicine given several physician-crafted prompts, emphasizing the need for critical evaluation of these LLMs to prevent racial biases. BiasMedQA \cite{schmidgall2024addressing} assess cognitive biases of LLMs in medical QA tasks. In addition to demographic bias, \citet{Jin2023BetterTA} further explores the language bias in medical QA tasks. \citet{Zack2024}, \citet{yang2024unmasking} and \citet{poulain2024bias} analyze specific bias of LLMs in clinical tasks with specific human-annotated references that limit to a small scope of analysis (e.g., a limited number of diagnoses). All these works also lack an evaluation of intrinsic bias. Although a recent work \cite{cross-care} evaluates intrinsic bias between diagnosis and demographics, the number of evaluated diagnoses is still limited to the requirement of reference data and does not provide a corresponding evaluation of extrinsic bias.

\section{\benchmark}
\label{sec:clinicalbias}

In this section, we first introduce the general setups of \benchmark grounded in a medical diagnosis setting. We then introduce the bias evaluations from both the intrinsic and extrinsic perspectives where the former examines disparities in LLM's representative associations between diagnoses and demographics, while the latter assesses performance differences when demographic information is varied within the given context of patient data. Figure~\ref{fig:pipeline} provides an illustration for \benchmark.

\label{sec: task_data_setup}

\paragraph{Diagnosis Decision Space} We focus on the bias associated and reflected in LLMs' diagnostic capability, where the diagnoses are identified by the International Classification of Diseases, Tenth Revision, Clinical Modification (ICD-10-CM) codes~\cite{icd-10-cm}. These well-known codes are adopted by numerous healthcare providers to catalog all medical diagnoses. It also categorizes diagnoses from high (Level 1)  to low (Level 5) levels:  L1 - Chapters; L2 - Blocks; L3 - Categories; L4 - Sub-categories; L5 - Full code. At L5, there are 94,739 diagnoses in total.

\paragraph{Demographic Attributes} In \benchmark, we select several attributes for studies, including demographics such as sex (i.e., Female and Male) and ethnicity (White, Black, Hispanic, and Asian). Additionally, we also study ``insurance types'' (Medicaid, Medicare, Other), an attribute that is unique to clinical settings.

\paragraph{Demographic-specific Diagnosis}
\label{sec:demo_diagnosis}

In the medical domain, it is an important and essential issue to recognize that certain diagnoses are specific to particular demographics. This specificity arises from correct causal factors that affect different groups in varying ways. Properly acknowledging this distinction ensures accurate and fair clinical assessments, thereby improving healthcare outcomes for all demographic groups. We utilize the female-only\footnote{\url{https://www.icd10data.com/ICD10CM/Codes/Rules/Female_Diagnosis_Codes}} and male-only\footnote{\url{https://www.icd10data.com/ICD10CM/Codes/Rules/Male_Diagnosis_Codes}} codes defined in  ICD-10-CM  to identify sex-specific diagnoses, resulting in 3,116 and 529 diagnoses respectively. Conversely, identifying ethnicity-specific diagnoses is challenging because the absence of robust clinical evidence linking specific diagnoses to particular ethnic groups, without resorting to stereotypes or problematic assumptions, complicates the establishment of such categories \cite{Omiye2023}. Therefore, we leave it as future work and only focus on the sex-specific diagnoses in this work.

\subsection{Evaluation of Intrinsic Bias}
\label{sec:implicit_eval}
Intrinsic bias studies the inherent associations between diagnosis and demographics in LLMs. 
In psychological research, Implicit Association Test (IAT) is a popular tool used to reveal the unconscious bias of humans by measuring the strength of subjects' automatic associations between concepts and targets \cite{iat}. In the context of NLP, IAT has been adapted to assess the intrinsic bias of language models through representation associations between {\tt target words} (e.g., names of different genders) and {\tt attribute words} (e.g., terms related to careers or families) that are embedded in models' internal representation \cite{weat, ceat, Bai2024MeasuringIB}. Our evaluation of intrinsic bias is also based on IAT with several important modifications.

To adapt IAT for our diagnostic task, we use diagnoses and demographic groups as attributes and targets respectively. We obtain the diagnoses from ICD-10-CM, and adopt people's names as a proxy for different demographics following existing works~\cite{sandoval-etal-2023-rose, an-etal-2023-sodapop, iat, weat}. These works show using names as a demographic proxy is advantageous because names inherently reflect demographic attributes like ethnicity and sex, allowing us to analyze biases without relying on explicit demographic labels, which may make the model {\it conscious} about the specific bias we are investigating (i.e., contradicted to {\it unconsciousness} of intrinsic bias).

Following \citet{sandoval-etal-2023-rose} and \citet{an-etal-2023-sodapop}, we collect the top 5 most popular names by {\it sex (Female and Male) and ethnicity (White non-Hispanic, Black non-Hispanic, Hispanic and Asian and Pacific Islander)}~\footnote{We call them as ``White'', ''Black'', ''Hispanic'' and ''Asian'' for simplicity, which is on a par with extrinsic part.} from the ``Popular Baby Names'' dataset~\footnote{\url{https://data.cityofnewyork.us/Health/Popular-Baby-Names/25th-nujf}}
, which provides more fine-grained ethnicity groups than traditional IAT that only provides White and Black.

\paragraph{Quantification of the Intrinsic Bias}
\label{sec:assocmad}
Previous works with bias evaluation based on IAT typically adopt the metric that is designed to measure the difference in association strengths between  \textit{two target groups} (e.g., male and female) with attribute words (e.g., career and family)~\cite{weat}. However, our diagnosis task involves a total of 8 demographic groups. Thus, we introduce a novel metric,  Association-disparity Metric (\mbox{AssocMAD}), which adapts the Mean Absolute Deviation (MAD) to measure disparities in associations across \textit{multiple} target groups.

In brief, MAD measures the average absolute distance between each data point and the mean, providing a robust estimate of disparity, which is defined by Equation~\ref{eq:mad}. Here \( x_i \) is the observed value, \( \mu \) is the mean of these values, and \( n \) is the total number of observations. 

\begin{equation}
    \label{eq:mad}
    \text{MAD} = \frac{1}{n} \sum_{i=1}^n |x_i - \mu|
\end{equation}

In embedding-based IATs such as WEAT \cite{weat} and CEAT \cite{ceat}, association scores are calculated using embedding distances (e.g., cosine similarity) between attributes and targets that are not applicable in recent LLMs. Instead, for decoder-only LLMs that are designed for autoregressive generation, we adopt a common alternative method -- {\it probability-based bias evaluation} \cite{cross-care, dong2023probing, nangia2020crows} that use next-token probability for the association calculation. 
Specifically,  we adopt the prompt ``{\tt \{$d_i$\} is related to the name:}'' because it clearly instructs an LLM to assign names based on diagnosis, while keeping the model {\it unconscious} about the specific bias we are investigating (i.e., ensure ``unconsciousness'' of intrinsic bias). We then measure the probability of each name ($n_i$) appearing as the next token for the association score corresponding to diagnosis $d_i$. For names tokenized into multiple sub-tokens, we calculate the joint probability of all sub-tokens. With the aforementioned approach, given a set of diagnoses $D=\{d_1,d_2,d_3,\ldots\}$ and a set of demographic groups $G=\{g_1,g_2,g_3,\ldots\}$, the association score $s\left(d_i, g_j\right)$ is defined as:
\begin{equation}
    \label{eq:score_func}
    s\left(d_i, g_j\right)=\frac{1}{\left|W_{g_j}\right|} \sum_{k=1}^{\left|W_{g_j}\right|} p\left( w_k^{g_j} | d_i\right)
\end{equation}
where $W_{g_j}=\{w_1^{g_j},w_2^{g_j},w_3^{g_j},\ldots\}$ is the set of stimuli (e.g., name) in group $g_j$, and $p\left( w_k^{g_j} | d_i\right)$ is the probability of $w_k^{g_j}$ being the next token conditioned on $d_i$.
Since the clinical task contains several levels of diagnoses, when evaluating at level $L$, the association score is calculated as an aggregation over diagnoses at the finest (L5) level: $s(d_i^L, g_j) = \sum_{l=1}^L s(d_i^{l_5}, g_j)$, where $d_i^l$ are codes at L5 level corresponding to $d_i^L$. 
For each diagnosis $d_i$, the $\text{\mbox{AssocMAD}}_{d_i}$ measure the association disparities of $d_i$ across all groups and is calculated as:
\begin{equation}
    \label{eq:mad_di}
    \text{\mbox{AssocMAD}}_{d_i}=\frac{1}{|G|} \sum_{j=1}^{|G|}\left|s\left(d_i, g_j\right)-\mu_{s\left(d_i\right)}\right|
\end{equation}
where $\mu_{s\left(d_i\right)}$ is the mean of the association scores across all groups $G$. As $s\left(d_i, g_j\right)$ might be extremely small and also affects term $\left|s\left(d_i, g_j\right)-\mu_{s\left(d_i\right)}\right|$, making the MAD less insightful. By normalizing this term by dividing by $\mu_{s\left(d_i\right)}$, we adjust the $\text{\mbox{AssocMAD}}_{d_i}$ to a relative scale that better highlights the disparity.

\begin{equation}
    \label{eq:mad_di_normalized}
    \text{\mbox{AssocMAD}}_{d_i}=\frac{1}{|G|} \sum_{j=1}^{|G|}\left|\frac{s\left(d_i, g_j\right)-\mu_{s\left(d_i\right)}}{\mu_{s\left(d_i\right)}}\right|
\end{equation}

The final \mbox{AssocMAD} is defined as the macro mean of the $\text{\mbox{AssocMAD}}_{d_i}$ across all considered diagnoses, which indicates disparities in overall diagnosis associations across all demographic groups. 

\paragraph{Single-demographic \mbox{AssocMAD}} In the original \mbox{AssocMAD} calculation, the mean association score \(\mu_{s\left(d_i\right)}\) is averaged across all demographic groups (i.e., White Male, $\cdots$, Asian Female), providing an overall measure of disparity without highlighting the disparities specific to either sex or ethnicity. We introduce a variation of \mbox{AssocMAD} for a specific demographic. Specifically, we first average the association scores for all names within that demographic group. For instance, to assess sex-specific disparities, we average the association scores of all female names as the score for females (i.e., incl. female across White, Black, Hispanic and Asian). The demographic-specific \mbox{AssocMAD} for each diagnosis \(d_i\) can be revised as follows where $x$ denotes the specific demographic type (e.g., sex).
\begin{equation}
    \label{eq:mad_di_x}
    \begin{multlined}
        \text{Single-demographic AssocMAD}_{d_i,x} \qquad\qquad \\
        = \frac{1}{|G_{\text{x}}|} \sum_{j=1}^{|G_{\text{x}}|} \left| \frac{s_{\text{x}}\left(d_i, g_j^{\text{x}}\right) - \mu_{s_{\text{x}}\left(d_i\right)}}{\mu_{s_{\text{x}}\left(d_i\right)}} \right| \qquad\qquad
    \end{multlined}
\end{equation}
\paragraph{Correctness of Sex Preference} \mbox{AssocMAD} indicates disparities among groups, rather than a preference towards a specific group. To quantify whether a model can prefer the correct sex for sex-specific diagnoses (i.e., aligned with medical knowledge), we introduce the ratio $\frac{\text{\# correct preference}}{\text{\# (fe)male-only diagnoses}}$ as
 \mbox{\it correctness of sex preference}, where ``correct preference'' for female-only diagnoses is defined as association score (same as Equation~\ref{eq:score_func}) of female is larger than male and vice versa for male-only diagnoses.

\subsection{Evaluation of Extrinsic Bias}
\label{sec:explict_eval}

Extrinsic bias aims to reflect the disparity of models' performance on end tasks towards various demographics. To set up a controlled setting and reflect the influence of the bias attribute only, we define extrinsic bias as the change in model's performance when we switch the demographic information in a clinical note where we use the diagnosis task introduced in {\sc CliBench} \cite{clibench} as the end task. 

\textsc{CliBench} is the first benchmark to evaluate the diagnostic performance of recent LLMs using a large output space (i.e., ICD-10-CM). This setting makes the performance sensitive to input perturbations, allowing for a more precise examination of potential biases. It contains representative clinical tasks induced from the Medical Information Mart for Intensive Care (MIMIC)-IV database, which is comprised of de-identified Electronic Health Records (EHR) for patients admitted to the Beth Israel Deaconess Medical Center.
Unlike previous medical QA datasets, the samples in {\sc CliBench} are clinical admission record in real-world with detailed clinical notes, demographic information, and a large diagnosis decision space (i.e., ICD-10-CM). This allows us to reveal LLMs' extrinsic biases in a realistic clinical setting. 

From the evaluation set of {\sc CliBench}, which consists of clinical cases balanced across chapters of diagnosis, service departments, and care units, we use all 199 instances with ``white'' ethnicity, ``male'' sex, and ``other'' insurance type \footnote{``Medicare'' and ``Medicaid'' are insurance programs that aim at helping minorities. ``Other'' covers a more privileged group of people, which is more suitable to serve as a baseline in our experiment.} (i.e., most privileged group) as the default diagnosis samples for counterfactual intervention. 
In this evaluation set, 23.6\% of the samples contain at least one sex-specific diagnosis. We consider these samples as \textit{sex-specific} and others as \textit{sex-neutral}. We follow the same prompt template introduced in {\sc CliBench} including the demographic information and clinical notes. In Appendix~\ref{sec:eval_stat}, we provide data statistics of this curated evaluation set compared to the original one in {\sc CliBench} and the prompt template. 

To build the counterfactual example, we replace the demographic information with a corresponding group. When the demographic information is about sex, we also replace pronouns and gendered honorifics in the instance to ensure the context remains coherent and accurate. We define \textbf{extrinsic bias score} as the \textit{difference in LLM's performance on the diagnosis prediction task}. In real-world clinical settings, we prioritize reducing Type 2 errors to avoid missing diagnoses. To achieve this, we use the difference in recall rate averaged on diagnostic abstraction levels while inputting default and counterfactual patient information to the models. To mitigate the potential effect of positional bias of demographic information, we computed the average score of putting demographic information before and after clinical notes. For the diagnosis prediction task, we follow {\sc CliBench} to extract valid diagnosis predictions from natural language output generated by LLMs and convert the natural language to ICD code whose description has the highest cosine similarity with the text.

\begin{table*}[!ht]
\centering
\small
\begin{tabular}{{l}{l}{r}*{5}{c}{c}}
\toprule
\#  & Model/Method          & Params   & L1    & L2    & L3    & L4    & L5    & Avg \\ \midrule
1 & Mistral Instruct v0.1  & 7B    & 0.15                       & 0.19                       & 0.22                       & 0.36                       & 0.55                       & 0.29                       \\
2 & Mistral Instruct v0.2  & 7B    & \underline{0.12}  & \underline{0.14}  & \underline{0.18}  & {0.24}              & {0.33}              & \underline{0.20}  \\
3 & Mistral Instruct v0.3  & 7B    & 0.26                       & 0.36                       & 0.41                       & 0.45                       & 0.59                       & 0.41                       \\
4 & BioMistral DARE        & 7B    & 0.18                       & 0.19                       & 0.25                       & 0.39                       & 0.57                       & 0.32                       \\
5 & Mistral Instruct v0.1  & 46.7B & {0.32}              & {0.43}              & {0.54}              & {0.63}              & {0.79}              & {0.54}              \\
\midrule
6 & LLaMA2 Chat            & 7B    & 0.72                       & 0.73                       & 0.75                       & 0.80                       & 0.85                       & 0.77                       \\
7 & Meditron               & 7B    & \underline{0.12}  & {0.16}              & {0.20}              & \underline{0.23}  & \underline{0.30}  & \underline{0.20}  \\
8 & Asclepius              & 7B    & 0.70                       & 0.68                       & 0.67                       & 0.66                       & 0.65                       & 0.67                       \\
9 & LLaMA2 Chat            & 13B   & \textbf{0.88}  & \textbf{0.91}  & \textbf{0.96}  & \textbf{1.00}  & \textbf{1.09}  & \textbf{0.97}  \\
10 & LLaMA2 Chat            & 70B   & 0.38                       & 0.41                       & 0.45                       & 0.49                       & 0.52                       & 0.45                       \\
\midrule
11 & LLaMA3 Instruct        & 8B    & {0.34}              & {0.36}              & {0.37}              & {0.37}              & {0.38}              & {0.36}              \\
12 & OpenBioLLM             & 8B    & {0.54}              & {0.55}              & {0.55}              & {0.53}              & {0.58}              & {0.55}              \\
13 & LLaMA3 Instruct        & 70B   & 0.47  & 0.49 & 0.50  & 0.50  & 0.51  & 0.49  \\
\bottomrule
\end{tabular}
\caption{Overall intrinsic bias. 
A higher value means more bias. The column ``Avg.'' is \mbox{AssocMAD} averaged across all diagnosis group levels (from L1 to L5). The smallest and largest values of each granularity are in \underline{underline} and \textbf{bold}, respectively. 
}
\label{tab:main_open}
\end{table*}

\section{\benchmark Experiments}

\subsection{LLMs for the Evaluation} 

We assess LLMs from two well-known open-source families: Mistral \cite{jiang2023mistral} and LLaMA \cite{touvron2023llama}. We include three versions of Mistral Instruct 7B (v0.1 to v0.3) and a mixture-of-experts version of v0.1, Mistral Instruct v0.1 (8x7B). For the LLaMA family, we include LLaMA 2 Chat (7B, 13B, and 70B)  and the more recent LLaMA3 Instruct (8B and 70B). We also consider their medically adapted versions: BioMistral DARE \cite{labrak2024biomistral}, which is obtained by first pre-training  Mistral Instruct v0.1 on PubMed Central corpus and then merging it with the original Mistral Instruct v0.1 for better generalizability. Meditron \cite{chen2023meditron} continues to pre-train LLaMA2 7B on PubMed articles \cite{kweon2023publicly}, while Asclepius finetunes LLaMA2 7B through QA synthesized from PMCPatients case reports \cite{zhao2022pmc}. Additionally, OpenBioLLM \cite{OpenBioLLMs} tailors the LLaMA3 8B model to specialized medical instruction and ranking datasets.

\subsection{Results of Intrinsic Bias}
In this section, we show intrinsic bias in LLMs for sex-neutral and sex-specific diagnoses respectively. For the former, LLMs are expected to show a close to zero AssocMAD score. For the latter, LLMs are expected to show the correct correlation for sex. 

\subsubsection{Sex-neutral Diagnoses}
Table~\ref{tab:main_open} reports the intrinsic bias for sex-neutral diagnoses in terms of \mbox{AssocMAD}, which measure the disparity across all demographic groups. A smaller value indicates a smaller disparity (i.e., less biased). We summarize several key observations from this scenario: 
\textit{I. Overall, all models show smaller \mbox{AssocMAD} at a higher level}, which can be attributed  to the very broad coverage of diagnoses in coarser analyses. On average, Mistral Instruct v0.2 7B and Meditron 7B (row \#2 and \#7) perform the best with the lowest \mbox{AssocMAD}. Mistral Instruct v0.2 7B achieves the best at higher levels (L1 to L3), whereas Meditron 7B achieves the best at lower levels (L4 to L5). LLaMA2 Chat 13B (\#9) performs the worst across all levels, indicating the most biased. 
\textit{II. Interestingly, larger models such as Mistral Instruct v0.1 46.7B (\#5), LLaMA2 Chat 13B (\#9), and LLaMA3 Instruct 70B (\#13) did not outperform their smaller 7B versions (\#1, 6 and 11)}, which suggests that simply increasing model size does not necessarily mitigate bias and may exacerbate it due to the model’s complexity and the varied data it is exposed to during training.
For the most up-to-date version of Mistral and LLaMA, Mistral v0.3 7B (\#3) is more biased than LLaMA3 Instruct 8B (\#11) with an averaged \mbox{AssocMAD} of 0.41, compared to 0.36. 
\textit{III.  Among medically adapted models, Meditron (7B) stands out, as introduced earlier, with the lowest \mbox{AssocMAD}}. However, with the same based model (i.e., LLaMA2 Chat 7B; \#6), Asclepius 7B (\#8) achieves slighter improvement. A possible reason may be that its fine-tuning data (i.e., PMC-Patients case reports) contain more relations between the diagnosis codes and patients' demographics. 
\textit{IV. Models pre-trained or fine-tuned on medical corpora with incorrect association may even introduce more intrinsic bias.} For example, both BioMistral DARE 7B (\#4) and OpenBioLLM 8B (\#12) are  worse than their base models Mistral Instruct v0.1 7B (\#1) and LLaMA3 Instruct 8B (\#11), respectively.

We compute \mbox{AssocMAD} over the entire set of real-world diagnoses (i.e., ICD-10-CM), rather than relying on a sample set. Therefore, there is no need for significance testing related to diagnosis sampling. To analyze the sensitivity to name selection, we further compute \mbox{AssocMAD} via top-K name resampling by frequency, where $k=\{1,2,3,4\}$ and results in Table~\ref{tab:main_open} are the most certain with $k=5$ (highest overall frequency). In summary, we find the above key observations remain consistent across all resamplings. We report the detailed results in the Appendix~\ref{sec:name_resample}.

To highlight the disparities in sex or ethnicity separately as introduced in Section~\ref{sec:assocmad}, we report the single-demographic \mbox{AssocMAD} (see Equation~\ref{eq:mad_di_x}) of sex and ethnicity in Table~\ref{tab:main_single}. Although there should be no association difference between females and males for sex-neutral diagnoses, all models still exhibit sex bias (i.e., non-zero \mbox{AssocMAD} in column ``Sex''). \mbox{AssocMAD} for both sex and ethnicity reveal similar bias trends on par with previous overall analysis. Mistral Instruct v0.2 7B and Meditron 7B continue to show minimal bias. Mistral Instruct v0.1 46.7B and LLaMA3 Instruct 70B are still more biased to their smaller versions  in both sex and ethnicity. Interestingly, although LLaMA2 Chat 13B is more biased than 7B   in terms of sex, they are the same biased in ethnicity with the same \mbox{AssocMAD} of 0.64. LLaMA3 Instruct  still shows better bias control compared to its predecessor in both sex and ethnicity than Mistral Instruct v0.3.

\begin{table}[!t]
\centering
\small
\begin{tabular}{{l}{l}{r}{r}{r}}
\toprule
\#  & Model/Method          & Params   & Sex & Ethnicity  \\ \midrule
1 &  Mistral Instruct v0.1   &  7B     &  0.19                        &   0.23          \\
2 &  Mistral Instruct v0.2   &  7B     &  0.14                        &  \underline{0.14}  \\
3 &  Mistral Instruct v0.3   &  7B     &  0.32                        &   0.19          \\
4 &  BioMistral DARE         &  7B     &  0.21                        &   0.24          \\
5 &  Mistral Instruct v0.1   &  46.7B  &  0.36                        &   0.39          \\
\midrule
6 &  LLaMA2 Chat             &  7B     &  0.39                        &   \textbf{0.64}  \\
7 &  Meditron                &  7B     &  \underline{0.12}               &   \underline{0.14}  \\
8 &  Asclepius               &  7B     &  0.39                        &   0.50                       \\
9 &  LLaMA2 Chat             &  13B    &  \textbf{0.53}            &   \textbf{0.64}  \\
10 &  LLaMA2 Chat            &  70B    &  0.22                        &   0.28                       \\
\midrule
11 &  LLaMA3 Instruct         &  8B     &  0.17                       &   0.19              \\
12 &  OpenBioLLM              &  8B     &  0.19                       &   0.45              \\
13 &  LLaMA3 Instruct         &  70B    &  0.25                       &   0.26          \\
\bottomrule
\end{tabular}
\caption{Averaged \mbox{single-demographic AssocMAD} for sex and ethnicity across all diagnosis levels.}
\label{tab:main_single}
\end{table}

\subsubsection{Sex-specific Diagnoses}
\label{sec:sex_intrinsic_stat}

We report correctness of sex preference for sex-specific diagnoses in Table~\ref{tab:sex_specific_intrinsic}. 
Although some models demonstrate high correctness for either female-only or male-only diagnoses, they often fail to exhibit a balanced and reasonable sex preference for both. For instance, LLaMA2 Chat 13B shows a high correctness of 0.94 for female-only diagnoses but a low correctness of 0.14 for male-only diagnoses. This indicates it still prefers females even for male-only diagnoses. Similarly, LLaMA3 Instruct 8B shows a striking imbalance with correctness of 0.80 for female-only and 0.03 for male-only diagnoses. Although this seems to be counter-intuitive, a possible reason is their association differences are usually quite small, which suggesting model just struggles to identify these as sex-specific rather than counter-intuitively exhibiting a pronounced opposite-sex preference. This trend is also shown in the medically adapted models and is even more severe than the corresponding base version (e.g., Asclepius compared to LLaMA2 Chat 7B), which suggests that the medically adapted training may not help them learn the correct knowledge of sex preference for these sex-specific diagnoses. 

\begin{table}[!htbp]
\centering
\small
\begin{tabular}{{l}{l}{r}{r}{r}}
\toprule
\#  & Model/Method          & Params   & Female & Male  \\ \midrule
1 & Mistral Instruct v0.1    & 7B & 0.13 & 0.81 \\
2 & Mistral Instruct v0.2    & 7B & 0.16 & 0.81 \\
3 & Mistral Instruct v0.3    & 7B & 0.63 & 0.53 \\
4 & BioMistral DARE          & 7B & 0.13 & 0.81 \\
5 & Mistral Instruct v0.1    & 46.7B & 0.19 & 0.81 \\
 \midrule
6 & LLaMA2 Chat              & 7B & 0.00 & 0.99 \\
7 & Meditron                 & 7B & 0.26 & 0.74 \\
8 & Asclepius                & 7B & 0.00 & \textbf{1.00} \\
9 & LLaMA2 Chat              & 13B & \textbf{0.94} & 0.14 \\
10 & LLaMA2 Chat              & 70B & 0.85 & 0.11 \\
 \midrule
11 & LLaMA3 Instruct          & 8B & 0.80 & 0.03 \\
12 & OpenBioLLM               & 8B & 0.23 & 0.54 \\
13 & LLaMA3 Instruct          & 70B & 0.70 & 0.08 \\
 \bottomrule
\end{tabular}
\caption{Correctness of sex preference for sex-specific diagnoses, the larger the better. The columns ``Female''/``Male'' denote female/male-only diagnoses. The largest correctness are highlighted in \textbf{bold}.}
\label{tab:sex_specific_intrinsic}
\spacemagic{\vspace{-1em}}
\end{table}

\begin{table*}[!ht]
\small
\centering
\begin{adjustbox}{max width=\textwidth}
  \begin{tabular}{llrc*{5}{c}{c}}
    \toprule
    \multirow{2}{*}{\#} & \multirow{2}{*}{Model/Method} & \multirow{2}{*}{Params}
    & \multicolumn{1}{c|}{Sex} 
    & \multicolumn{3}{c|}{Ethnicity} 
    & \multicolumn{2}{c}{Insurance} 
    \\ \cline{4-9}
    & & & Female & Black & Hispanic & Asian & Medicaid & Medicare  \\ \midrule
    1 & Mistral Instruct v0.1  & 7B       & \cellcolor{red!10}{-3.53 (-15.26\%)} & \cellcolor{red!10}{-0.24 (-1.04\%)} & \cellcolor{green!10}{0.07 (0.28\%)} & \cellcolor{green!10}{0.36 (1.53\%)} & \cellcolor{red!10}{-0.61 (-2.63\%)} & \cellcolor{red!10}{-0.90 (-3.89\%)}  \\
    2 & Mistral Instruct v0.2  & 7B       &\cellcolor{red!10}{-2.74 (-7.97\%)} & \cellcolor{red!10}{-0.58 (-1.67\%)} & \cellcolor{red!10}{-0.56 (-1.65\%)} & \cellcolor{red!10}{-0.81 (-2.36\%)} & \cellcolor{red!10}{-0.30 (-0.89\%)} & \cellcolor{red!10}{-0.10 (-0.31\%)}  \\ 
    3 & Mistral Instruct v0.3  & 7B       &\cellcolor{red!10}{-1.24 (-3.84\%)} & \cellcolor{red!10}{0.20 (0.60\%)} & \cellcolor{red!10}{-0.17 (-0.52\%)} & \cellcolor{red!10}{-0.31 (-0.97\%)} & \cellcolor{green!10}{0.12 (0.37\%)} & \cellcolor{red!10}{-0.58 (-1.80\%)} \\ 
    4 & BioMistral DARE        & 7B       &\cellcolor{red!10}{-4.30 (-21.15\%)} & \cellcolor{red!10}{-0.41 (-2.04\%)} & \cellcolor{red!10}{-0.36 (-1.77\%)} & \cellcolor{red!10}{-0.46 (-2.26\%)} & \cellcolor{green!10}{1.61 (7.89\%)} & \cellcolor{green!10}{0.27 (1.35\%)} \\ 
    5 & Mixtral Instruct v0.1  & 46.7B    & \cellcolor{red!10}{-0.38 (-1.09\%)} & \cellcolor{green!10}{0.17 (0.49\%)} & \cellcolor{red!10}{-0.30 (-0.85\%)} & \cellcolor{red!10}{-0.11 (-0.32\%)} & \cellcolor{red!10}{-0.76 (-2.18\%)} & \cellcolor{green!10}{0.31 (0.89\%)} \\
    \midrule
    6 & LLaMA2 Chat            & 7B       & \cellcolor{red!10}{-0.36 (-1.38\%)} & \cellcolor{green!10}{0.25 (0.94\%)} & \cellcolor{red!10}{-0.45 (-1.68\%)} & \cellcolor{green!10}{0.01 (0.02\%)} & \cellcolor{green!10}{0.01 (0.04\%)} & \cellcolor{red!10}{-0.41 (-1.54\%)} \\
    7 & Meditron               & 7B       &\cellcolor{green!10}{0.54 (39.14\%)} & \cellcolor{green!10}{0.38 (27.08\%)} & \cellcolor{green!10}{0.38 (27.08\%)} & \cellcolor{green!10}{0.38 (27.08\%)} & \cellcolor{green!10}{0.45 (32.13\%)} & \cellcolor{green!10}{0.49 (35.74\%)} \\
    8 & Asclepius              & 7B       &\cellcolor{red!10}{-2.14 (-63.88\%)} & \cellcolor{red!10}{-2.09 (-62.54\%)} & \cellcolor{red!10}{-2.01 (-60.15\%)} & \cellcolor{red!10}{-2.04 (-60.90\%)} & \cellcolor{green!10}{1.54 (45.97\%)} & \cellcolor{green!10}{1.91 (57.01\%)} \\
    9 & LLaMA2 Chat            & 13B      & \cellcolor{red!10}{-0.59 (-2.08\%)} & \cellcolor{green!10}{0.52 (1.83\%)} & \cellcolor{red!10}{-0.16 (-0.54\%)} & \cellcolor{green!10}{0.02 (0.09\%)} & \cellcolor{red!10}{-0.12 (-0.40\%)} & \cellcolor{green!10}{0.27 (0.96\%)}\\
    10 & LLaMA2 Chat            & 70B      & \cellcolor{red!10}{-0.32 (-1.05\%)} & \cellcolor{red!10}{-0.25 (-0.81\%)} & \cellcolor{red!10}{-0.31 (-1.00\%)} & \cellcolor{red!10}{-0.70 (-2.24\%)} & \cellcolor{green!10}{0.46 (1.50\%)} & \cellcolor{green!10}{0.14 (0.47\%)} \\
    \midrule
    11 & LLaMA3 Instruct        & 8B       & \cellcolor{red!10}{-1.05 (-3.05\%)} & \cellcolor{red!10}{-0.61 (-1.78\%)} & \cellcolor{red!10}{-0.46 (-1.34\%)} & \cellcolor{red!10}{-0.56 (-1.64\%)} & \cellcolor{red!10}{-0.28 (-0.80\%)} & \cellcolor{red!10}{-0.10 (-0.28\%)}\\
    12 & OpenBioLLM             & 8B       &\cellcolor{red!10}{-4.90 (-95.34\%)} & \cellcolor{red!10}{-5.11 (-99.32\%)} & \cellcolor{red!10}{-5.11 (-99.42\%)} & \cellcolor{red!10}{-5.10 (-99.13\%)} & \cellcolor{green!10}{1.09 (21.09\%)} & \cellcolor{green!10}{1.48 (28.67\%)} \\
    13 & LLaMA3 Instruct        & 70B      &\cellcolor{green!10}{0.39 (0.82\%)} & \cellcolor{green!10}{0.03 (0.07\%)} & \cellcolor{green!10}{0.19 (0.41\%)} & \cellcolor{green!10}{0.30 (0.64\%)} & \cellcolor{green!10}{0.08 (0.16\%)} & \cellcolor{green!10}{0.09 (0.18\%)} \\
    \bottomrule
  \end{tabular}
\end{adjustbox}
\caption{Extrinsic bias score of models when separately replacing sex, ethnicity and insurance. Numbers are increased (\colorbox{green!10}{$+$}) or decreased (\colorbox{red!10}{$-$}) recall averaged across all diagnosis levels compared to the origin (see Table~\ref{tab:ext_origin}). Zero numbers are expected for column ``Insurance'', as it is not a causal factor of diagnosis compared to sex and ethnicity. The percentage represents the change as a proportion of the original recall rate.}
\label{tab:counter_res}
\end{table*}

\subsection{Results of Extrinsic Bias}
We first analyze the overall performance of the models to evaluate how sensitive the models are to counterfactual demographic information. Then, we compare samples with sex-specific diagnoses with other samples to further analyze where the performance change comes from. 

\subsubsection{Overall Result}

We expect a model could be slightly sensitive to sex and ethnicity as they are potential causal factors to diagnosis, but non-sensitive to insurance.
From Table~\ref{tab:counter_res}, we observe:
\textit{I. Decreased diagnosis performance for female instances across models.} Comparing models when given factual and counterfactual sex information, we observed a consistent performance drop in almost all models when we provide counterfactual female sex information. However, when given counterfactual ethnicity or insurance, the results are mixed. 
\textit{II. Medical adapted LLMs are more sensitive to demographic information change.} Comparing the general domain models and their corresponding medically adapted models, we observed a more significant performance change in medically adapted ones, indicating they are more sensitive to the counterfactual demographic information given in the prompt.

We notice although the results from Tables~\ref{tab:counter_res} do not achieve statistical significance, our pioneering benchmark provides critical evidence of extrinsic clinical bias, highlighting the fairness challenges within the clinical domain. One possible reason for the lack of significance is our deliberate minimization of changes in counterfactual interventions (i.e., altering only a few words), which helps reduce the introduction of unnecessary noise and biased assumption. However, we acknowledge that future work building on our pioneering method, could explore modifying other medical information linked with demographics (e.g., body metrics). While this could offer new insights, it also risks making biased assumptions without strict medical support, potentially distorting the detection of genuine bias and introducing contradictions in patient cases.
\begin{table}[t!]
\centering
\small
\begin{adjustbox}{max width=\columnwidth}
\begin{tabular}{{l}{l}{r}{r}{r}}
\toprule
\#  & Model/Method          & Params   & Sex-neutral & Sex-specific  \\ \midrule
1 & Mistral Instruct v0.1    & 7B & \cellcolor{red!10}-3.93 (-16.61\%) & \cellcolor{red!10}-1.49 (-7.22\%) \\
2 & Mistral Instruct v0.2    & 7B & \cellcolor{red!10}-2.83 (-8.20\%) & \cellcolor{red!10}-2.24 (-6.72\%) \\
3 & Mistral Instruct v0.3    & 7B & \cellcolor{red!10}-1.16 (-3.56\%) & \cellcolor{red!10}-1.66 (-5.22\%) \\
4 & BioMistral DARE          & 7B & \cellcolor{red!10}-4.23 (-21.19\%) & \cellcolor{red!10}-4.58 (-20.56\%) \\
5 & Mistral Instruct v0.1    & 46.7B & \cellcolor{red!10}-0.14 (-0.41\%) & \cellcolor{red!10}-1.59 (-4.67\%) \\
 \midrule
6 & LLaMA2 Chat              & 7B & \cellcolor{red!10}-0.45 (-1.67\%) & \cellcolor{red!10}-0.33 (-1.30\%) \\
7 & Meditron                 & 7B & \cellcolor{green!10}0.50 ( 34.04\%) & \cellcolor{green!10}0.78 ( 77.19\%)\\
8 & Asclepius                & 7B & \cellcolor{red!10}-2.24 (-64.21\%) & \cellcolor{red!10}-1.59 (-61.38\%) \\
9 & LLaMA2 Chat              & 13B & \cellcolor{red!10}-0.73 (-2.55\%) & \cellcolor{green!10}0.11 ( 0.38\%) \\
10 & LLaMA2 Chat              & 70B &  \cellcolor{red!10}-0.29 (-0.93\%) & \cellcolor{red!10}-0.52 (-1.56\%) \\
 \midrule
11 & LLaMA3 Instruct          & 8B & \cellcolor{red!10}-0.91 (-2.65\%) & \cellcolor{red!10}-1.78 (-5.17\%) \\
12 & OpenBioLLM               & 8B & \cellcolor{red!10}-5.00 (-94.59\%) & \cellcolor{red!10}-4.38 (-100.00\%) \\
13 & LLaMA3 Instruct          & 70B & \cellcolor{green!10}0.22 (0.45\%) & \cellcolor{green!10}1.29 (2.73\%) \\
 \bottomrule
\end{tabular}
\end{adjustbox}
\caption{Extrinsic bias score separated into sex-neutral and sex-specific samples when only replacing sex.  Zero and negative numbers are expected for columns ``Sex-neutral'' and ``Sex-specific'' respectively.}
\label{tab:sex_specific_extrinsic}
\end{table}

\subsubsection{Sex-specific Samples}

The overall results in Table~\ref{tab:counter_res} column ``Sex'' aggregates results for both sex-specific and sex-neutral samples. However, a performance drop is expected only in sex-specific samples, as these contain at least one male-specific diagnosis, which would naturally lead to decreased performance when the sex is changed to female. In contrast, models should not be sensitive to sex for sex-neutral samples, and any performance difference in these samples indicates sex bias.

To analyze the breakdown contributor of the overall performance differences of sex, we divide our dataset into sex-specific and sex-neutral sample groups according to the criteria introduced in Section~\ref{sec: task_data_setup}. In Table~\ref{tab:sex_specific_extrinsic}, we separately report their extrinsic bias scores when only replacing sex. In sex-neutral samples, performance differences (i.e., sex bias) occur for all models. In sex-specific samples, several models (\# 7, 9 and 13) show counter-intuitive performance increase. 
We observe this is because the majority of sex-neutral diagnoses within these samples contribute more positively than the negative impact of the male-specific diagnoses. However, the positive changes of sex-neutral diagnoses are still biased as we expect no change for them when perturbing sex. For all medically adapted models, their performance differences are even larger than their based counterparts in sex-neutral samples (i.e., more biased), which suggests that their medically adapted training may not help to identify these samples are sex-neutral, therefore unable to avoid sex bias when predicting diagnoses.

\section{Conclusion}

In conclusion, \benchmark\ serves as a pioneering benchmark specifically crafted to concurrently and systematically evaluate both intrinsic and extrinsic bias in LLMs applied for clinical decision-making tasks. Notably, we introduce a novel metric, \mbox{AccocMAD}, allowing for a more comprehensive assessment of disparities across multiple demographic groups, and several advanced strategies to adapt counterfactual intervention into the evaluation of extrinsic bias in clinical setting. Our extensive experiments across a range of popular and medically adapted LLM models, particularly from the Mistral and LLaMA families, unveil prevalent biased behaviors, with even some medical adapted LLMs performing worse than their base counterparts and lacking effective bias mitigation. This work not only underscores the critical need for bias mitigation in medical adapted LLM applications but also sets a new standard for future bias evaluations of AI systems in healthcare.

\section*{Limitations}

In this first version of \benchmark, we focus on the most representative clinical task -- diagnosis task. Future works can further adapt \benchmark to other clinical tasks such as procedures, lab test orders and prescriptions. Although we have included key factors (i.e., sex, ethnicity and insurance) that may lead to clinical bias, future works can continue to explore other factors such as age, marital status, and so on. In addition to utilizing MIMIC-IV dataset for the evaluation of extrinsic bias, other clinical datasets can also be developed for bias evaluation. Furthermore, more advanced techniques of bias mitigation can be developed and evaluated utilizing \benchmark. Another future direction could be studying bias issues in medical or clinical datasets to understand how the bias comes about. 

\section*{Ethical Considerations}

The bias evaluation results we reported are highly related to the selected patient clinical cases. The bias observed for models is only a reflection of the model's prediction on a particular dataset using a particular definition of bias. This paper focuses on the bias defined as the disparity of the model's diagnostic understanding and capabilities. We would like to raise the warning that the bias we refer to in this work does not represent the overall bias in society.

The bias attributes examined in this work are limited by the data availability (e.g., patient profile information available in EHR) and disease categorization evidence (e.g., definition of sex-specific diagnoses). For selected bias attributes, there are other potential biases for demographic groups that are not covered in this work. There are also more bias attributes to be considered in real-world applications, beyond sex, ethnicity and insurance types.

The analysis performed in this work reflects an aggregated bias level for a specific demographic group at a certain diagnostic granularity; the bias exhibited by the model could be more significant than the aggregated level reported in this paper. We would like to raise awareness that domain experts' manual examination is needed, especially when applying the proposed bias analysis method in specific instances.

\bibliography{climb_bib,revised_arxiv}

\appendix

\section{Prompt Template and Evaluation Set}
\label{sec:eval_stat}

Table~\ref{tab:prompt_template} show the our prompt template for extrinsic bias evaluation which is the same as {\sc CliBench}.
This prompt of diagnsis task is realistical for clinical bias evaluation because 1) It’s realistically used by clinical resource management units for resource forecasting, billing departments for auditing and insurance companies for risk reduction; 2) Infering diagnosis could indirectly benefit patient treatment, such as medication recommendation for online pharmacy and ordering lab tests; 3) It examines a comprehensive set of clinical capabilities of the LM covering a long chain of clinical operations, including clinical note understadning, diagnosis decision space understanding, reasoning with complex information, and clinical knowledge about facts about lab test and drugs, and associations between sympothms and diagnosis.

\begin{table}[!ht]
    \small
    \begin{tcolorbox}
    {\bf System Prompt} \\
    You are a professional clinician in a hospital with expert knowledge in medical and clinical domains. The task is to make a list of diagnoses for this patient based on the provided information of the patient. The diagnosis can be in ICD-10-CM code format (such as S12.000G), or natural language description of the disease. Separate each diagnosis with a new line. Please provide as many diagnoses as you can until you are not confident about your diagnosis decision. \\ \\
    {\bf User Prompt} \\
    Patient information: sex is {\textbraceleft\textbraceright}, ethnicity is {\textbraceleft\textbraceright} ...
    {\textbraceleft Clinical Note ... \textbraceright}
    What are the diagnoses for this patient?
    \end{tcolorbox}
    \caption{Prompt template of the diagnosis task in \sc{CliBench}}
    \label{tab:prompt_template}
\end{table}

\begin{figure}[!ht]
    \centering
    \includegraphics[width=\columnwidth]{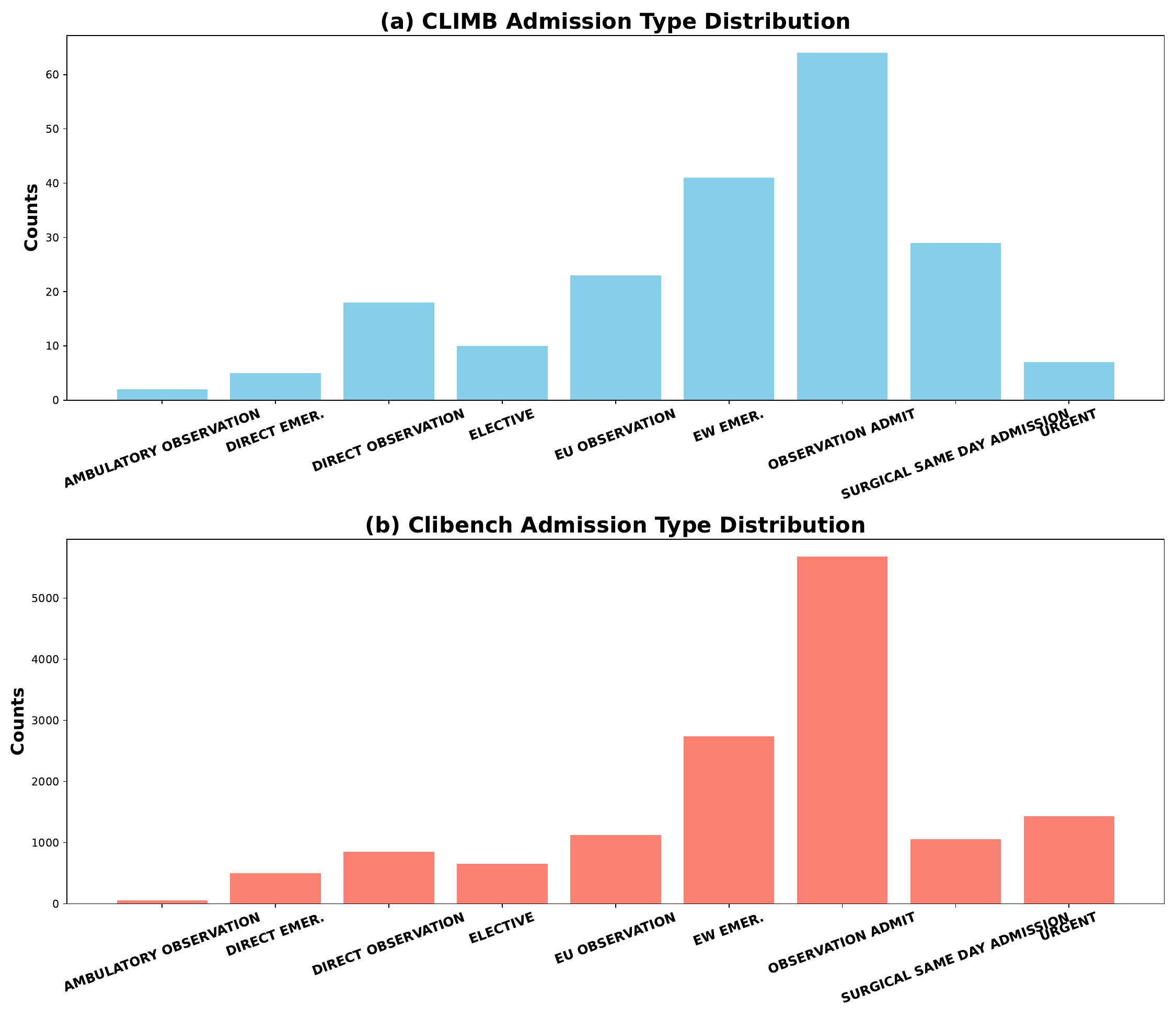}
    \caption{Admission distribution comparison.}  
    \label{fig:adm_dist}
\end{figure}

\begin{figure}[!ht]
    \centering
    \includegraphics[width=\columnwidth]{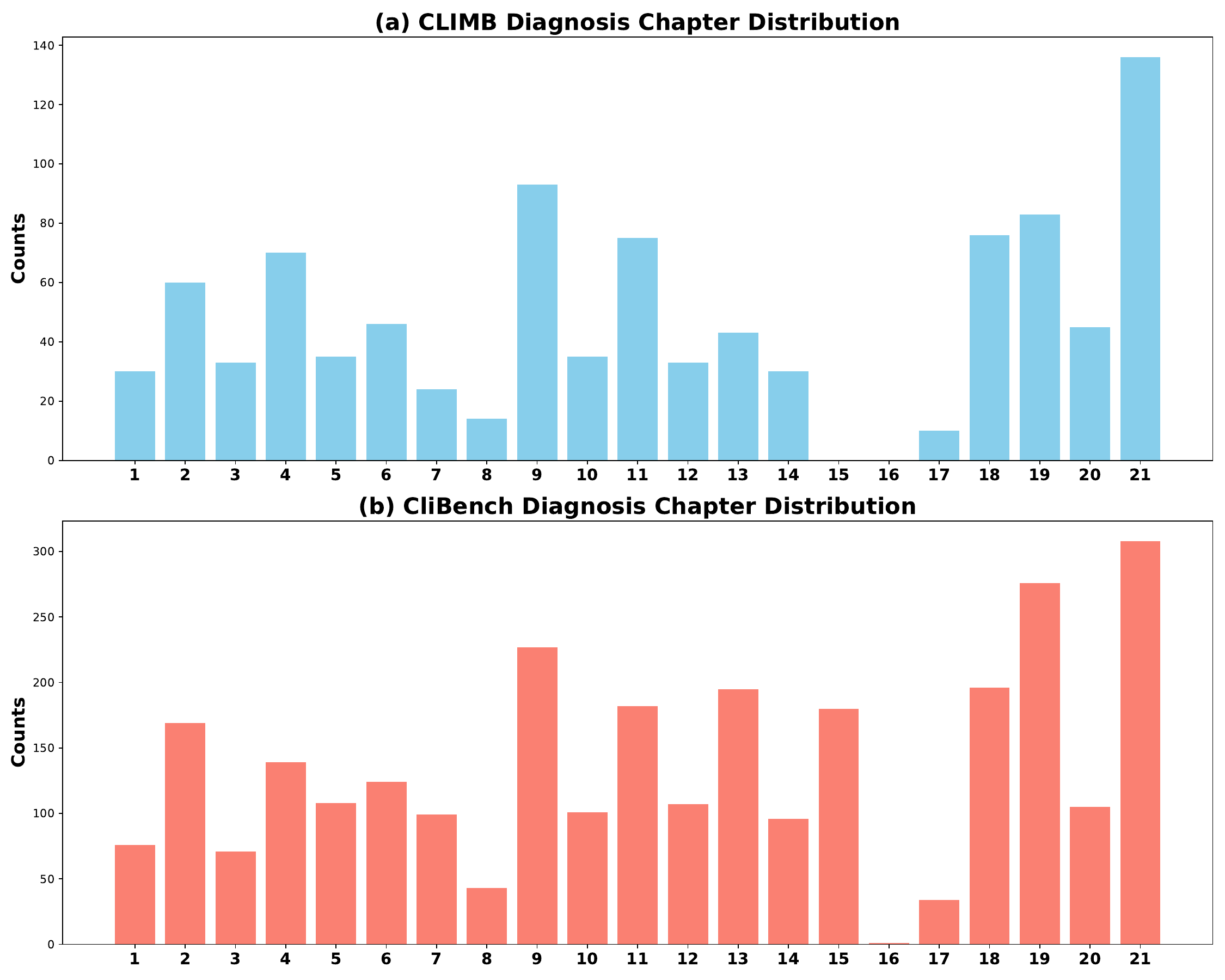}
    \caption{Diagnosis distribution comparison.}  
    \label{fig:diag_dist}
\end{figure}

Figure~\ref{fig:adm_dist} and Figure~\ref{fig:diag_dist} show the statistics of CLIMB evaluation set. Comparing sub-figures (a) and (b), it is evident that our subset maintains the sample diversity of the {\sc CliBench} dataset.

\section{Original Performance of Evaluation Set for Extrinsic Bias}
\label{sec:origin}

We show the performance of the original evaluation set for extrinsic bias in Table~\ref{tab:ext_origin}.

\begin{table}[!ht]
\centering
\small
\begin{tabular}{{l}|{l}|{c}|{c}}
\toprule
\#  & Model/Method          & Params   & Origin  \\ \midrule
1 & Mistral Instruct v0.1    & 7B & 23.16  \\
2 & Mistral Instruct v0.2    & 7B & 34.35 \\
3 & Mistral Instruct v0.3    & 7B & 32.41  \\
4 & BioMistral DARE          & 7B & 20.34\\
5 & Mistral Instruct v0.1    & 46.7B & 34.81  \\
 \midrule
6 & LLaMA2 Chat              & 7B & 26.54  \\
7 & Meditron                 & 7B & 1.38 \\
8 & Asclepius                & 7B & 3.35  \\
9 & LLaMA2 Chat              & 13B & 28.66 \\
10 & LLaMA2 Chat              & 70B & 31.02 \\
 \midrule
11 & LLaMA3 Instruct          & 8B & 34.25  \\
12 & OpenBioLLM               & 8B & 5.15 \\
13 & LLaMA3 Instruct          & 70B & 47.80  \\
 \bottomrule
\end{tabular}
\caption{Numbers are the recall rate averaged across all diagnosis levels, the larger the better.}
\label{tab:ext_origin}
\end{table}

\section{Name Resampling for Intrinsic Bias}
\label{sec:name_resample}

Table~\ref{tab:main_open_var_k} shows the results of different top-K name resampling. We observe that our key observations from Table~\ref{tab:main_open} ($k=5$) remain consistent across all resamplings.

\begin{table*}[!ht]
\centering
\small
\begin{tabular}{{l}{l}{r}*{5}{c}{c}}
\toprule
\#  & Model/Method          & Params   & L1    & L2    & L3    & L4    & L5    & Avg \\ \midrule
$k=1$ & & & & & & & & \\
& Mistral Instruct v0.1  & 7B    & 0.35 & 0.41 & 0.49 & 0.69 & 0.92 & 0.57 \\
& Mistral Instruct v0.2  & 7B    & 0.33 & 0.39 & 0.48 & 0.65 & 0.84 & 0.54 \\
& Mistral Instruct v0.3  & 7B    & 0.57 & 0.68 & 0.72 & 0.83 & 1.01 & 0.76 \\
& BioMistral DARE        & 7B    & 0.40 & 0.44 & 0.53 & 0.72 & 0.92 & 0.60 \\
& Mistral Instruct v0.1  & 46.7B & 0.73 & 0.83 & 0.97 & \textbf{1.08}    & \textbf{1.24}              & 0.97 \\
\midrule
& LLaMA2 Chat            & 7B    & 0.74 & 0.75 & 0.77 & 0.77 & 0.79 & 0.76 \\
& Meditron               & 7B    & \underline{0.22} & \underline{0.27} & \underline{0.34} & \underline{0.40} & \underline{0.50} & \underline{0.35}           \\
& Asclepius              & 7B    & 0.97 & 0.95 & 0.95 & 0.98 & 0.98 & 0.97 \\
& LLaMA2 Chat            & 13B   & \textbf{1.11}    & \textbf{1.10}    & \textbf{1.10}    & \textbf{1.08}              & 1.13 & \textbf{1.10}              \\
& LLaMA2 Chat            & 70B   & 0.81 & 0.83 & 0.84 & 0.88 & 0.87 & 0.85 \\
\midrule
& LLaMA3 Instruct        & 8B    & 0.53 & 0.55 & 0.56 & 0.58 & 0.61 & 0.57 \\
& OpenBioLLM             & 8B    & 0.96 & 0.95 & 0.93 & 0.94 & 0.99 & 0.95 \\
& LLaMA3 Instruct        & 70B   & 0.88 & 0.89 & 0.88 & 0.84 & 0.83 & 0.86 \\
\midrule
\midrule
$k=2$ & & & & & & & & \\
& Mistral Instruct v0.1  & 7B    & 0.24 & 0.29 & 0.36 & 0.53 & 0.74 & 0.43 \\
& Mistral Instruct v0.2  & 7B    & 0.23 & 0.27 & 0.34 & 0.47 & 0.62 & 0.39 \\
& Mistral Instruct v0.3  & 7B    & 0.38 & 0.50 & 0.56 & 0.64 & 0.82 & 0.58 \\
& BioMistral DARE        & 7B    & 0.28 & 0.32 & 0.39 & 0.56 & 0.74 & 0.46 \\
& Mistral Instruct v0.1  & 46.7B & 0.54 & 0.66 & 0.80 & 0.90 & \textbf{1.04}              & 0.79 \\
\midrule
& LLaMA2 Chat            & 7B    & 0.50 & 0.53 & 0.56 & 0.56 & 0.59 & 0.55 \\
& Meditron               & 7B    & \underline{0.17} & \underline{0.20} & \underline{0.26} & \underline{0.30} & \underline{0.39} & \underline{0.26}           \\
& Asclepius              & 7B    & 0.78 & 0.76 & 0.75 & 0.77 & 0.78 & 0.77 \\
& LLaMA2 Chat            & 13B   & \textbf{0.93}    & \textbf{0.93}    & \textbf{0.96}    & \textbf{0.95}              & 1.02 & \textbf{0.96}              \\
& LLaMA2 Chat            & 70B   & 0.56 & 0.59 & 0.62 & 0.64 & 0.65 & 0.61 \\
\midrule
& LLaMA3 Instruct        & 8B    & 0.56 & 0.56 & 0.56 & 0.57 & 0.57 & 0.56 \\
& OpenBioLLM             & 8B    & 0.73 & 0.74 & 0.73 & 0.72 & 0.78 & 0.74 \\
& LLaMA3 Instruct        & 70B   & 0.73 & 0.75 & 0.75 & 0.71 & 0.70 & 0.73 \\
\midrule
\midrule
$k=3$ & & & & & & & & \\
& Mistral Instruct v0.1  & 7B    & 0.18 & 0.23 & 0.29 & 0.44 & 0.64 & 0.36 \\
& Mistral Instruct v0.2  & 7B    & 0.16 & 0.20 & 0.26 & 0.36 & 0.48 & 0.29 \\
& Mistral Instruct v0.3  & 7B    & 0.29 & 0.42 & 0.48 & 0.55 & 0.70 & 0.49 \\
& BioMistral DARE        & 7B    & 0.21 & 0.24 & 0.31 & 0.47 & 0.65 & 0.38 \\
& Mistral Instruct v0.1  & 46.7B & 0.43 & 0.55 & 0.68 & 0.79 & 0.93 & 0.68 \\
\midrule
& LLaMA2 Chat            & 7B    & \textbf{0.84}    & \textbf{0.84}    & \textbf{0.84}    & \textbf{0.89}              & 0.92 & \textbf{0.87}              \\
& Meditron               & 7B    & \underline{0.14} & \underline{0.18} & \underline{0.22} & \underline{0.26} & \underline{0.34} & \underline{0.23}           \\
& Asclepius              & 7B    & 0.76 & 0.74 & 0.73 & 0.71 & 0.70 & 0.73 \\
& LLaMA2 Chat            & 13B   & 0.78 & 0.77 & 0.81 & 0.84 & \textbf{0.96}              & 0.83 \\
& LLaMA2 Chat            & 70B   & 0.39 & 0.44 & 0.47 & 0.50 & 0.54 & 0.47 \\
\midrule
& LLaMA3 Instruct        & 8B    & 0.42 & 0.44 & 0.45 & 0.45 & 0.45 & 0.44 \\
& OpenBioLLM             & 8B    & 0.61 & 0.62 & 0.61 & 0.59 & 0.64 & 0.61 \\
& LLaMA3 Instruct        & 70B   & 0.58 & 0.60 & 0.61 & 0.61 & 0.61 & 0.60 \\
\midrule
\midrule
$k=4$ & & & & & & & & \\
& Mistral Instruct v0.1  & 7B    & 0.16 & 0.20 & 0.25 & 0.39 & 0.59 & 0.32 \\
& Mistral Instruct v0.2  & 7B    & 0.14 & 0.17 & 0.22 & 0.30 & 0.42 & 0.25 \\
& Mistral Instruct v0.3  & 7B    & 0.27 & 0.38 & 0.43 & 0.48 & 0.62 & 0.44 \\
& BioMistral DARE        & 7B    & 0.19 & 0.21 & 0.28 & 0.43 & 0.61 & 0.34 \\
& Mistral Instruct v0.1  & 46.7B & 0.35 & 0.48 & 0.60 & 0.70 & 0.86 & 0.60 \\
\midrule
& LLaMA2 Chat            & 7B    & 0.78 & 0.79 & 0.80 & 0.85 & 0.88 & 0.82 \\
& Meditron               & 7B    & \underline{0.13} & \underline{0.16} & \underline{0.20} & \underline{0.24} & \underline{0.32} & \underline{0.21}           \\
& Asclepius              & 7B    & 0.73 & 0.71 & 0.70 & 0.68 & 0.67 & 0.70 \\
& LLaMA2 Chat            & 13B   & \textbf{0.90}    & \textbf{0.93}    & \textbf{0.98}    & \textbf{1.01}    & \textbf{1.10}    & \textbf{0.98}              \\
& LLaMA2 Chat            & 70B   & 0.39 & 0.43 & 0.47 & 0.51 & 0.54 & 0.47 \\
\midrule
& LLaMA3 Instruct        & 8B    & 0.37 & 0.39 & 0.40 & 0.40 & 0.40 & 0.39 \\
& OpenBioLLM             & 8B    & 0.55 & 0.56 & 0.56 & 0.54 & 0.59 & 0.56 \\
& LLaMA3 Instruct        & 70B   & 0.53 & 0.55 & 0.56 & 0.58 & 0.58 & 0.56 \\
\bottomrule
\end{tabular}
\caption{Results of top-K name resampling for overall intrinsic bias.}
\label{tab:main_open_var_k}
\end{table*}

\clearpage

\end{document}